# Assessing the Reusability of Public Speech Resources for Low-Resource Languages: A Central Kurdish Case Study

Hiwa Asadpour
University of Saarland and Goethe University Frankfurt
asadpour@lingua.uni-frankfurt.de

**Abstract**
Kurdish is spoken by millions of people, but very little technology can read it aloud. That changed recently: a team published three Kurdish computer voices, 35 hours of recorded speech, and a paper describing the work, all free to download. This review checks how well those public files match the paper that describes them. The research itself is careful about its limits. The files are not. A settings file lists equipment that was never used, the test recordings are left unlabeled among the training data, and a coding fault mishandles long numbers. The download page also claims a stronger result than the researchers do, and recommends one voice for general use. That last point matters most. Kurdish is not one language with one agreed way of writing it, and these voices were built from three people reading prepared texts. Every stage of the process strips out regional and everyday speech. English and German voices rest on centuries of dictionaries and description that let a wrong pronunciation be shown to be wrong; Kurdish has far less, so software decides instead, unchecked. The result sounds fluent, but it is one narrator's reading style, not the language. Almost all of this is fixable with information the team already holds, and none of the fixes would change a result. They would change who can check it. A missing detail is a nuisance to an engineer and a wall to the community linguists who work on the varieties with no voice at all. The license is the exception: whether the audiobook owners permit corrected versions to be shared decides whether the next Kurdish variety builds on this work or starts from nothing.

## 1. Introduction and background information

Despite its substantial speaker population, Central Kurdish remains poorly represented in publicly available speech technology resources. The language has a written standard based on a modified Arabic script, yet publicly documented and evaluated speech systems and datasets remain scarce. The related-work section of Mohammadamini, Shamsi, and Tahon's LREC 2026 paper (p. 664) illustrates this gap: a 21-hour Tacotron 2 system reported in 2024, for which neither the models nor the data were released; 13 hours of studio recordings from a separate project; and Meta's multilingual MMS models. The authors note these lack detailed evaluation, public datasets, and voice-cloning capability.

A recent release changes the situation, but it also raises a different question: how closely does the public release correspond to the research described in the paper? The release provides 35 hours and 20 minutes of transcribed speech from three speakers, three trained text-to-speech models, and published evaluation code. The paper reports MOS scores between 4.01 and 4.08 for the three systems and provides a level of documentation that is unusual among speech technology releases for Central Kurdish. This study does not independently assess audio quality; its scope is the documentation, evaluation design, licensing and released artifacts.

A researcher who reads the publication encounters one account of what was done. A developer who opens the Hugging Face page encounters a different and less complete one. A third party who downloads the files finds artifacts that do not quite correspond to either.

None of this indicates misconduct, and this study makes no such claim. Rather, it describes a gap between a well-documented piece of research and the materials through which most users will encounter it. That gap matters in a low-resource language, where there are few alternative datasets, models, or tools against which a release can be checked.

This study examines that gap across five areas: how the models were built, how they were evaluated, what can be reproduced, what the license permits, and what the release records about the language itself. Each section describes what the sources show, explains what follows from those findings, and proposes what would close the gap.
**On sources.** Everything below is checked against the full published paper (LREC 2026, pp. 664–673), the released model and dataset files, and the upstream components. Where the paper and the model card disagree, both are cited. Where an earlier assumption of mine is contradicted by the paper, I state that directly rather than leaving the assumption in place.

**2. What the paper establishes that the repository does not**

Three questions a developer would reasonably ask are answered clearly in the publication and nowhere in the release itself.
The first concerns how the models were made. Section 3 states plainly that the team fine-tuned what it calls the English checkpoint of F5-TTS-base, with a footnote linking to the exact starting weights (p. 666)[1]. This is not a minor detail. Building a high-quality Kurdish system from eleven hours of audiobooks and adapting a model pretrained on a large Chinese-and-English corpus with eleven hours of Kurdish are different achievements, and readers should be able to tell which one they are looking at. That distinction is available in the paper and lost on the model card. The card describes the models as “based on the F5-TTS architecture”. That wording points to a borrowed design rather than borrowed weights, and leaves a reader with the more impressive of the two readings.
The second concerns an oddity in the released files. Anyone who opens ‘vocab.txt’ finds 2,545 entries containing Mandarin pinyin syllables with tone numbers, Korean hangul, Japanese kana, Cyrillic, Greek, Hebrew, Thai and extended Latin, and nothing Kurdish-specific. Encountered without explanation, this looks like an error. The paper accounts for it: F5-TTS ordinarily avoids grapheme-to-phoneme conversion,

> *“however its input vocabulary miss many Kurdish characters”*,

so the pipeline converts Kurdish text to phonemes first,

> *“to make it compatible with the vocabulary of the F5-TTS baseline model”* (p. 666).

The reasoning is sound, and the paper states it directly. F5-TTS was selected against stated design priorities of

> *“data efficiency, speaker flexibility, and robustness to noisy or non-standardized data”* (p. 665),

properties that depend on starting from pretrained weights, and therefore on keeping the vocabulary those weights were trained against. The authors also observe that

> *“Central Kurdish language is almost a phonemic language and have very high precision Grapheme-to-Phoneme (G2P) tools”* (p. 666),

so routing text through phonemes costs little and solves the character-coverage problem without retraining anything. The paper is explicit that the conversion serves the vocabulary rather than the language: it is applied

> *“to make it compatible with the vocabulary of the F5-TTS baseline model”* (p. 666).

The trade-off is that the pipeline now depends on a component the models cannot see. The G2P converter’s decisions are baked into the training data, and the paper’s own citation for that component, Mahmudi and Veisi (2021), reports that Kurdish orthography remains contested, particularly around the short vowel */i/* and the letters ی and و. Those unresolved cases are resolved anyway, by software, before the model sees anything. Text that bypasses the converter, or that it handles poorly, arrives at a vocabulary with no Kurdish characters in it. The number-handling bug described later in this study is precisely that failure.
The release rests on a design choice documented in the paper and nowhere in the artifacts. Anyone working from the repository alone cannot see why the pipeline is built as it is.
The third concerns the license, and it matters because the restriction looks like a choice when it is not. Both the models and the dataset carry CC BY-NC-ND 4.0: no commercial use, no modified versions. That is unusually restrictive for a research release and would ordinarily invite the question of why the authors selected it. Section 8 answers that:

> *"the writers and narrators of the audiobooks granted permission for the use of their data under the CC BY-NC-ND 4.0 license"* (p. 671).

---

[1] The paper describes the starting point as the English checkpoint of F5-TTS-base and links to SWivid/F5-TTS/tree/main/F5TTS_Base. The upstream repository documents that checkpoint as trained on Emilia 95K zh&en. Chinese and English rather than English alone. The distinction matters for how the low-resource result should be read: the amount and diversity of pretraining data determines how much eleven hours of Kurdish had to accomplish. A second, smaller mismatch sits alongside it: the paper’s footnote points to F5TTS_Base, while the configuration file in the model repository is named F5TTS_v1_Base.yaml, referring to the later version. Which of the two was used is not stated.

The no-derivatives restriction came from the rights holders as a condition of access. The researchers did not add it, and any criticism suggesting otherwise would be misdirected.

Recognizing the constraint does not dissolve its effects, and those effects fall on reproducibility rather than on commerce. The no-derivatives term reaches further than the non-commercial one, because verification in machine learning is largely a matter of producing modified versions: retraining from the same starting point to confirm the reported numbers, ablating a preprocessing step to test its contribution, fine-tuning on held-out speakers to check that a result generalizes. Each of those produces an adapted model, and publishing one is what makes the check visible to anyone else. A term that permits the work but not its redistribution allows verification in private and forbids it in public. That is close to forbidding it outright. An unreproduced finding and an unpublished reproduction carry the same weight in the literature.

The accountability question runs alongside it. Where a model cannot be corrected and redistributed, errors accumulate in place. The number-handling bug described later in this study is a small illustration: it is a two-line fix, and no one outside the project can publish it. Anyone who hits the number bug will hit it again in the next release.

**Toward better practice.** Two conventions would help. The first is to separate licenses by artifact rather than applying one term across a release. Recordings, transcripts, model weights and code carry different rights and different risks, and a permission covering audiobook content need not govern derived weights. Where rights holders are approached at the outset, asking specifically whether derived models may be redistributed, while the recordings themselves stay restricted, costs nothing and preserves the community's ability to build on the result.

The second is to state license provenance plainly: which terms are inherited, from which upstream source, and which were chosen. This release inherits non-commercial terms from the F5-TTS pretrained weights and no-derivatives from the audiobook permissions, and neither fact appears in the model card. Recording it tells a reader not only what they may do, but which party they would have to persuade to do more.

What connects these three cases is that the information exists and simply does not travel with the artifacts. The paper is the authoritative record, but the model card is what appears in search results, what a developer skims before downloading, and what tools index. A release that documents itself only in the publication leaves most of its users under-informed.

**What would close this.** Each gap has a predictable failure attached to it. A reader who does not know the models were fine-tuned from a large pretrained multilingual checkpoint will overstate what eleven hours of Kurdish achieved, and may cite the release as evidence for a claim it does not support. A developer who opens vocab.txt without the phonemic explanation will conclude the release is broken, or will feed raw Kurdish script to a model that cannot read it. A team evaluating the license will read no-derivatives as the researchers' position and either abandon the work or approach the wrong party about relaxing it, when the parties who could actually relax it are the audiobook rights holders. Three sentences on the model card, drawn from the paper's own wording, would prevent all three.

### 3. Where the released configuration and the paper diverge

The model repository ships a training configuration file, F5TTS_v1_Base.yaml, which the model card never references. It disagrees with the paper in two respects.

The file specifies a batch size of 38,400 frames per GPU and carries a comment reading *"8 GPUs, 8 * 38400 = 307200."* The paper reports that fine-tuning was performed *"independently for three datasets, leveraging 1 GPU (RTX8000, 48GB) in 2 days"* (p. 666). The file also specifies eleven epochs, while Figure 1 plots learning curves extending to roughly 275,000 training steps and the accompanying text discusses convergence *"during* *the first 100k iterations"* (p. 667). On eleven hours of audio, eleven epochs cannot approach 275,000 steps; the two figures differ by orders of magnitude.
The most likely explanation is prosaic. Configuration files of this kind are typically copied from an upstream template and edited only where necessary to launch a run. A diff against upstream confirms it: the dataset name was changed and the Weights & Biases logging block removed; the hardware comment and epoch count were not touched.
A machine-learning engineer will read the eight-GPU comment against the paper's one GPU, recognize a stale template, and adjust. The reader with most to gain from this release will not. Documentation and revitalization work on Kurdish varieties is largely carried out by linguists, teachers and community organizations, not by people who can tell an inherited default from a recorded setting. For that reader, the configuration is not a hint to be interpreted; it is the instruction manual. Following it means requesting eight GPUs for a job the authors ran on one, a request that in many institutions is refused, or costs a grant line that does not exist. Eleven epochs on a small corpus produces a model that has barely trained, and the reader has no way to tell whether the flat output reflects their data, their setup, or a number that was never meant to apply to them. The likely conclusion is that the method does not transfer, when in fact it does.
That misreading has a particular cost for Kurdish. The paper's own framing identifies Kurdish as a macrolanguage spanning Northern, Central, Southern, Hawrami, Zazaki and Laki (p. 664), while the release examined here covers Central Kurdish through three single-speaker systems. Extending it is the work the community needs, and the obstacle is not conceptual. The paper shows that eleven hours of audiobook recordings suffice when fine-tuning from a pretrained checkpoint, far below what studio recording implies. Whether other varieties have eleven hours available is not surveyed here. What stands between a Badini or Hawrami corpus and a working model is partly procedural: knowing which settings were actually used, on what hardware, for how long. Badini is a Northern Kurdish variety, written in Arabic script in Iraq, so extending to it involves a variety change and, depending on the source material, a script the Central Kurdish converter was not built for. Hawrami sits outside the Central Kurdish grouping altogether. Neither obstacle is conceptual, but neither is only a matter of copying settings. A configuration that misreports all three converts a reproducible recipe into a series of expensive guesses, and each failed attempt discourages the next.
Where several written standards coexist and none has institutional backing, the varieties that acquire speech technology first tend to keep the advantage, because subsequent tools are built and benchmarked against what already exists. Accurate configuration files are a small thing, but they determine how cheaply a second variety can follow the first.
A related and equally small problem: the paper's own footnotes direct readers to aranemini/tts4all-ckb-dataset and aranemini/ckb-f5-tts (p. 671). Both were renamed. Requested without credentials on 8 September 2026, each returned HTTP 307 redirecting to the current repositories, central-kurdish-tts4all and central-kurdish-tts. The published names still reach the artifacts; they no longer match the names in the repository URLs.
**What would close this.** Correct two lines in the configuration to match the reported run, or add a header noting which settings were used and which are inherited defaults. The redirect from the published names is already in place, so citations remain traceable.

### 4. What the evaluation supports, and what the model card claims

The evaluation is the most substantial part of the paper, and the clearest instance of the gap this study describes. The authors' claim is modest. The abstract describes the models as

> *"competitive with the models trained on dedicated TTS data recorded in the studio"* (p. 664).

The conclusion says audiobook data

> *"reach competitive results in comparison with high quality studio data"* (p. 670).

Section 4.2 describes the female audiobook result as *"marginally better"* (p. 668). The paper never claims one system beat another; it claims they perform similarly. The numbers support that.
The model card describes the same result differently. The female audiobook system, it says,

*"achieved the highest overall subjective score among the evaluated Central Kurdish TTS systems"*

and

*"provides highly natural speech synthesis suitable for general-purpose applications."*

That is a firmer claim than the researchers make, published on the page a developer reads before downloading. The study behind those numbers was substantial. Eighty-eight listeners produced 3,101 ratings across seven categories, answering

*"How close is the following voice to real human speech?"* on a scale from 1 to 5 (p. 668).

Each rated 40 sentences; each system was assessed on 140 sentences across seven categories, with listeners not informed that real speech was present. The results appear in Table 1:

Table 1: "Average and std of MOS per system"

| System | Natural speech | Synthetic | Std. dev. |
|---|---|---|---|
| audiobook-F | 4.42 | 4.08 | 0.98 |
| audiobook-M | 4.34 | 4.01 | 1.06 |
| studio-M | 4.40 | 4.06 | 1.04 |

"Table 2"in the source article, p. 668 and here as Table 1. Column headings as printed: speaker, natural, tts_mos, std_mos.

The ranking that the model card reports rests on 4.08 against 4.06. For scale, the per-category confidence intervals in Table 3 range from ±0.14 to ±0.19 on MOS (p. 669). The paper does not claim significance for a gap of two hundredths of a point. It calls the difference marginal, and that is the accurate description.

A further pattern in the table explains why the paper's cautious framing is the right one. Each system loses almost exactly the same amount relative to its own source recordings: 0.34, 0.33 and 0.34, differing by at most 0.01. The three systems also rank in the same order as the recordings they learned from. The paper observes as much:

*"those trained on data with higher MOS achieved higher MOS scores in the TTS system"* (p. 668).

Read one way, this is a useful result. If every system reproduces its speaker equally well, then cheap audiobook material did the job as effectively as purpose-recorded studio material, precisely the finding the paper claims, and precisely what a low-resource practitioner needs to know. Read another way, it means the ranking between the three tracks the quality of the source recordings rather than anything about the data source in itself.

The design produces that ambiguity. Each condition has a single speaker: one woman reading audiobooks, one man reading audiobooks, one man in a studio (Table 1, p. 665). *"Audiobook data"* therefore cannot be separated from the narrator's voice and skill, the recording environment, the material being read, or for the top-scoring system, the fact that it is the only female voice in the comparison. The limit is inherent in the available data, and the paper's language respects it. The model card's does not.

The design also fixes what the release can be used for. Three speakers is not a shortcoming of this study, the paper says so directly, noting that

*"the number of speakers is clearly not enough to train a Central Kurdish multi-speaker TTS model"* (p. 666).

Three separate mono-speaker systems were built for that reason. But the constraint follows the models downstream. Central Kurdish is "spoken by approximately 8 million people" (p. 664), and the release offers two male voices and one female. Any application built on it inherits that ratio: a screen reader, an announcement system, an educational tool for children will speak in one of three voices, two of them male.

The single female voice carries additional weight for a second reason. It is the highest-scoring system, so it is the one the model card recommends and the one a developer is most likely to select. Whatever is idiosyncratic about that speaker, her register, her pacing, the literary material she was reading, becomes the default sound of synthetic Kurdish for anyone who follows the recommendation. The paper's evaluation cannot distinguish which of her qualities the listeners were responding to, and neither can a user.

The card is not uniformly firmer than the paper. Its statement about objective signal quality is accurate. Section 4.1 sorts the objective metrics into naturalness, signal quality and intelligibility. On PESQ, the signal-quality measure, the studio system leads decisively: 4.16 on natural test and 4.00 in-domain, against 3.74 and 3.04 for the female audiobook system (pp. 666–669). The paper reports the same on page 668. The approximate figures the card gives elsewhere are, with one exception, the rounded means of the relevant Table 3 rows rather than imprecision: the studio system's predicted MOS is given as ~4.43 where the seven-category mean is 4.42.

**What would close this.** Align the model card's language with the paper's. *"Competitive with studio-recorded data"* and *"marginally better on subjective evaluation"* are both defensible and both already the authors' own words. Adding one line noting that each condition had a single speaker would let readers calibrate the comparison themselves.

**5. A measurement that depends on its own training data**

One issue in the evaluation is visible in the paper's own tables, and it affects how a widely-cited number should be read.

Character error rate measures intelligibility indirectly: run speech recognition over synthetic audio and count the divergence from the input text. The paper identifies the system used, SeamlessM4T V2 Large, fine-tuned for Central Kurdish, reported at 8% word error rate on the Asosoft benchmark and 20% on FLEURS (p. 667). The same fine-tuned system produced the training transcripts. Describing data preparation, the paper states that segmented utterances were

> *"transcribed automatically using a fine-tuned version of Seamless for Central Kurdish"*,

then manually revised (p. 665). The recognition system that wrote the text is the recognition system that scores the speech.

In many studies this would remain a theoretical concern. Here the effect is measurable, because "Table 3" reports character error rates for the *actual human recordings* used as controls, audio with no synthesis in it at all:

**Table 2:** ***"Averages and Confidence Intervals (95%) for metrics by training datasets and categories."***

| System | CER on real human speech |
|---|---|
| audiobook-F | 0.01 ± 0.01 |
| audiobook-M | 0.02 ± 0.01 |
| studio-M | 0.09 ± 0.08 |

Character error rate on actual human recordings, natural test category. Extracted from "Table 3", p. 669 from the original source.

The recognition system's error rate on real studio speech is 0.09, against 0.01 and 0.02 for the two audiobook speakers. The confidence interval on the studio figure is wide (±0.08), so the size of the gap is not well determined, but the direction is consistent and the gap exists before any text-to-speech system is evaluated. It reflects the measuring instrument, not the systems being measured.

The effect carries into the headline figures. The model card reports character error rates of 3.7%, 4.7% and 5.6%, the category means from "Table 3". The studio system's higher figure comes from two rows: natural test (0.09) and in-domain (0.08). These are the categories closest to its own recording domain, and against the female audiobook system they more than account for the whole gap in the seven-category mean. On categories furthest from any training domain it performs well: 0.02 on semantically unpredictable sentences and 0.03 on expressions, against 0.02 and 0.04 for the female audiobook system.

All three systems were measured with the same instrument, so the internal ranking retains meaning. But reading the aggregate figure as a straightforward statement about intelligibility overstates what it establishes, and the paper's assertion that character error rate *"is less than 8% and remains consistent across the three speakers"* (p. 668) does not quite match Table 3, where one row reads 0.09.

The consequences reach past this paper, because character error rate is the number that travels. It is a single figure, it is comparable across studies in a way that listening tests are not, and it is what a meta-analysis or a survey of low-resource TTS will extract. A reader who lifts 3.7%, 4.7% and 5.6% into a comparison table carries the instrument’s domain preference along with them, invisibly, and any conclusion drawn about audiobook versus studio data inherits it.

This circularity also limits what the metric can detect. The dependency runs the other way from what the paper’s structure suggests: the recognition system generated the transcripts that, after manual revision, became the training text, so the models learned to produce speech matching text that system had already shaped. Whether the recognition system was itself exposed to audiobook-domain Kurdish is not reported, and the paper cites it to a prior publication. A synthetic error that matches those patterns is more likely to be transcribed as intended than flagged as a mistake. The measurement is therefore least sensitive where the training data is thickest. For a language whose orthography admits competing spellings, that is not a small blind spot: a system could consistently realize an ambiguous vowel one way and the metric would register nothing, because the recognizer was trained to expect the same choice.

There is a practical cost for anyone extending the method. The paper presents a reusable recipe, and the evaluation pipeline is part of it, published as tts4all_eval. A researcher applying it to Badini or Hawrami may face the same loop: the recognition system used for evaluation could partly measure how closely the new variety resembles the Central Kurdish read speech on which the recognizer was developed. Without the control figures to compare against, they will have no way to tell how much of their result is the model and how much is the recognizer.

**What would close this.** Report the natural-speech control rates alongside the synthetic ones, or report each system’s error rate relative to its own natural baseline. Both are computable from data already in the paper, and either would let readers separate instrument behavior from system performance. Where possible, a second recognition system trained independently of the corpus would strengthen the intelligibility claim considerably. More generally, when a recognition system is used both to prepare training data and to evaluate output, saying so at the point where the metric is reported, rather than in a separate section on data preparation, lets any downstream reader weigh the number correctly without reconstructing the pipeline themselves.

**6. An undocumented preprocessing step**

The paper lists two preprocessing operations: normalization using the AsoSoft library, and grapheme-to-phoneme conversion (p. 666). The released inference script contains several more, and one of them does not function as intended.

Before normalization, the script replaces any unbroken run of nine or more digits with a placeholder:

‘‘‘python text = re.sub(r"\b\d{9,}\b", lambda m: f"<NUM:{m.group(0)}>", text)’’’

The purpose is clear enough: a twelve-digit number spelled out in full Kurdish words would be unmanageable, so the pipeline sets it aside. But nothing restores it. No reversing step exists anywhere in the file.

The placeholder therefore reaches the model as literal text, and one of its characters has no entry in the vocabulary at all: the punctuation block runs ‘;’, ‘=’, ‘>’, ‘?’, with no ‘<’ between them. The remaining characters are present, but only as symbols inherited from the English checkpoint that the Kurdish training data never exercised.

The failure is established by reading, not by listening. The substitution, the absence of any reversing step, and the missing < in vocab.txt are all directly checkable in the released files, and together they settle that unbroken numeric strings of nine digits or more are not processed as the code intends. What they do not settle is what the model produces in response, and that would require running the checkpoints under the license discussed in Section 7 which permits the test but not the publication of any adapted artifact from it. The reach is in any case limited: numbers written with spaces or separators pass through normally. The substitution originates in the research repository’s own inference script, and the third-party web demo inherited it unchanged.

The bug is narrow. The assumption behind it is not: the pipeline expects input to arrive in one script. The substitution operates on Western Arabic digits, and the AsoSoft normalizer that follows it converts Eastern Arabic numerals to a single form before number-to-word conversion runs. Anything outside that path is unhandled by design.

Kurdish supplies a great deal that falls outside it. Central Kurdish is written in modified Arabic script; Northern Kurdish is written in Latin in Turkey and Syria and in Arabic script in Iraq; and Kurdish in the former Soviet republics has historically been written in Cyrillic. Digits differ accordingly: ١٩٩٩ in Arabic-script contexts, 1999 in Latin ones. Suppose a Latin-script Kurmanji corpus reached this pipeline. Its characters would go to a converter built for Central Kurdish in Arabic script, which is the concrete form of the obstacle noted in Section 3. Whatever emerged would then meet a vocabulary assembled for other languages. Most of the Hawar Latin characters are in that vocabulary. Two are not: ḧ and ẍ, the converter's renderings of ح and غ. The failure mode would likely be silent rather than an error, though this study did not test it.

The paper is explicit that its scope is Central Kurdish. The caution is for the reuse it invites. The pipeline's script and orthographic assumptions are embedded in the AsoSoft dependency rather than stated in the code, so a linguist adapting the recipe to another variety inherits them without being told they exist. The single unreversed substitution is a visible instance of a general property: this preprocessing chain fails quietly rather than loudly.

The demo has a smaller issue that illustrates a general trap. It estimates required processing time from the character count of the input text but the model consumes phonemes, produced after numbers are expanded into words. A short sentence dense with figures expands considerably while its character count stays low, and the estimate falls short.

**What would close this.** For the first: restore the placeholder after normalization, or remove the substitution and let the numeral converter handle long numbers directly. For the second: estimate from the phoneme string. The code already computes it a few lines earlier. More usefully for anyone adapting the pipeline, a validation step that reports characters absent from the vocabulary before synthesis would convert every failure of this class from silent to visible, and costs a single lookup.

## 7. What can and cannot be reproduced

The audio, transcripts, checkpoints, vocabulary, configuration and inference script are all downloadable. The publication documents the architecture, the starting checkpoint, the hardware, the preprocessing chain, and the evaluation metrics with confidence intervals throughout.

The findings below come from retrieving each artifact and inspecting the released code in September 2026. Where retrieval failed, that is recorded as a property of the attempt rather than treated as evidence about the artifact.

The evaluation code is released as TTS4ALL_Eval on the LIUM GitLab, cited at p. 667; the repository carries the GNU AGPL v3.0. It was retrieved, installed and executed. Running the bundled test configuration over the four sample utterances reproduced the reference values shipped in tests/sample_audio/results.csv across all seven metrics, agreeing exactly on the deterministic measures and to six or seven significant figures on those computed on GPU. The package implements metrics across five aspects — intelligibility, naturalness, signal quality, speaker similarity and prosody — where the paper describes three (pp. 666–667). The released tool is broader than its published account. That is a point in its favor: a reader of the paper alone would not know that speaker-similarity and prosodic measures are available.

Reading the code also shows where the intelligibility measurement stops. The recognizer is selected by matching a language code from configuration against seven hardcoded branches in asr.py, covering Tunisian Arabic, Malay, Spanish, French, Bengali and Kabyle, together with a generic branch for w2v-BERT CTC models. None covers Kurdish, and both shipped example configurations set asr to false with the language left at French, so the intelligibility path is disabled by default in the released examples.

The generic branch is the relevant one. It loads a Wav2Vec2BertForCTC head with a SeamlessM4T feature extractor, the standard way of serving a fine-tuned SeamlessM4T encoder for recognition, and the same family as the model the paper describes at p. 667. It accepts an arbitrary checkpoint path from configuration and is set to load from local files only. It is a slot built for exactly this kind of model, left empty.

The checkpoint that would fill it is not distributed with the code. The paper cites its recognizer to prior work and gives no download location. The same Hugging Face account publishes a Central Kurdish ASR model whose card is empty; a request for its file listing, made without credentials, returned HTTP 401 with the header WWW-Authenticate: Bearer realm="Authentication required", charset="UTF-8". Its relationship to the model used in the paper is not documented. The evaluation infrastructure is therefore public while the component needed to use it for Kurdish is not, and the intelligibility column of Table 3 cannot be recomputed from the released materials.

One qualification is due, and it favors the authors. The recipe for building such a recognizer is published even though the checkpoint is not. evaluation/ASR/FT_W2V2Bert.py fine-tunes a Wav2Vec2BertForCTC model from facebook/w2v-bert-2.0 using a SeamlessM4T feature extractor, and its default data path names the AsoSoft Central Kurdish corpus. This places the gap in a different category from the missing evaluation sentences: the method is followable by anyone with comparable data, rather than closed.

Two further properties of the package affect anyone trying to run it. metrics.py imports fairseq and utmosv2 at module level, so no metric can be computed unless both are installed, even though most metrics use neither; and the quality-proxy metric fetches its checkpoint at runtime from a separate Hugging Face repository rather than shipping it. The release ships model definitions and training code rather than weights: the evaluation/directory contains an AASIST implementation, a quality-proxy network, and fine-tuning and inference scripts for Whisper and w2v-BERT CTC models, whiledata/audioanddata/textare empty. TheLIUM/TTSEval repository resolves publicly and holds three quality-proxy checkpoints of 414 MB each, so that dependency is satisfied. The bundled smoke test comprises four French utterances with expected values for signal quality, naturalness and quality-proxy metrics, but no character error rate, consistent with both example configurations disabling the intelligibility path. Together these mean the code is reproducible in the sense that it can be read and modified while running it depends on artifacts hosted elsewhere under separate access conditions.

The remaining gaps are narrow individually. Together they fall into a pattern: what is missing is the material an experienced engineer would supply from habit, and a linguist or community researcher would not know was needed.

The first concerns the test split. The paper states that "for each dataset 500 samples separated as the test set and the rest of the data is reserved for training the model" (p. 666). Table 1 lists dataset sizes of 6,044, 5,572 and 6,055 utterances, totaling 17,671, exactly the figure on the dataset card. The released corpus ships as a single train split containing all of them. The 1,500 held-out samples are present but unmarked. Anyone fine-tuning on this corpus will train on the paper's evaluation data.

The second concerns the listening test. The paper describes its seven categories in real detail (pp. 667–668): sentences containing English loanwords written in Kurdish script, emotionally expressive sentences, Kurdish idioms, grammatical but semantically nonsensical sentences, neutral news sentences, in-domain test sentences, and genuine speech. None of these sentence sets are published, so the subjective evaluation cannot be repeated by anyone else.

Beyond those two, several things a first-time user would need are absent from both the paper and the repository. There is no training script for the TTS models, and the configuration file specifies settings but not the command that consumes them. F5-TTS expects a directory holding a pipe-delimited metadata.csv and a wavs/ folder. Its prepare_csv_wavs.py converts these into the raw.arrow and duration.json files the trainer reads. The corpus ships as a Hugging Face parquet dataset instead, so a user must write that conversion before training can begin. There is no statement of how the audio was prepared beyond the resampling note on page 665, so a researcher preparing a new corpus must guess at loudness normalization, silence trimming and segment length policy. And there is no record of which checkpoint among the roughly 275,000 steps shown in Figure 1 was released, though the paper is explicit that this choice matters, opening Section 4 with the observation that no standard metric determines when to stop training (p. 666). The three released checkpoints are the outcome of a judgement the paper describes as unsolved and does not report making.

None of this obstructs a machine-learning engineer, who will write a manifest converter, pick sensible audio defaults, and take the final checkpoint without deliberating. The difficulty is that the paper's stated contribution is a method for rapid development of TTS in low-resource settings, and the people that method is aimed at are frequently not engineers. For them the undocumented steps are not conveniences omitted but the parts of the process they cannot infer, and the release offers no way to distinguish a setting chosen deliberately from one left at a default.

A smaller discrepancy: the paper and dataset card both total 17,671 utterances, while the Hugging Face dataset viewer reports 17,738 rows.

**What would close this.** The split column is the single most consequential change available, because its absence causes silent harm: someone fine-tuning on this corpus to add a fourth speaker will train on the paper's evaluation data, report improved numbers, and have no way to discover the contamination. One field prevents that permanently.

Publishing the 140 evaluation sentences per system is next. They carry no third-party rights, being the authors' own compositions, and without them the listening test that supports the paper's central claim cannot be repeated by anyone.

For the recognizer, two options exist and either would suffice. Releasing the fine-tuned checkpoint without an access barrier would make the intelligibility results reproducible with the code already published. Failing that, adding a Kurdish branch to asr.py pointing at any publicly available Central Kurdish recognizer would at least let others measure intelligibility on the same footing, even if not identically to the paper.

The remaining items are smaller but compound for the intended audience. A training command with the exact invocation used, a note on the audio preparation applied, and one sentence on which checkpoint was selected and why would together turn a documented result into a followable procedure. For a paper offering a recipe for low-resource languages, that difference is most of its practical value — and the same group's multi-dialect benchmark, whose multi-dialect resources cover Central, Northern and Southern Kurdish, with Hawrami and Badini announced as forthcoming, is precisely where a followable procedure would be applied next.

## 8. The license and its practical reach

The release is governed by CC BY-NC-ND 4.0, and page 671 explains why: the audiobook writers and narrators granted permission under that specific license. The upstream picture points the same way. The F5-TTS code is MIT-license d, but the pretrained weights these models were fine-tuned from carry a non-commercial restriction. The AsoSoft library is MIT and adds nothing.

The restrictions are therefore inherited from two directions, and holding the researchers responsible for them would be a mistake. At the same time, the way those terms are presented matters. Documentation frameworks in machine learning have treated licensing as reportable metadata since 2019. The datasheets proposal of Gebru et al. (2021) includes a distribution section asking whether third parties have imposed intellectual-property restrictions on the data and under what license or terms of use a dataset will be distributed. The model card framework of Mitchell et al. (2019) lists license among the model details a card should record. Hugging Face implements this structurally rather than as guidance: models and datasets occupy separate repositories, each carrying its own license field in the card metadata, so a release of this kind already has two places where provenance could be stated and neither is used for it. This release inherits non-commercial terms from the F5-TTS pretrained weights and no-derivatives from the audiobook permissions, yet neither fact is surfaced in the model card or dataset card. Recording it would tell a reader not only what they may do, but which party they would have to persuade to do more.

The effect on downstream users is unchanged by that fact. A Kurdish developer can download the models and use them in research. What becomes legally uncertain is fine-tuning them on an additional speaker, correcting the numeric-input bug and publishing the fix, adapting a voice toward a different variety, or converting the weights for deployment on a phone. Whether any of these constitutes a "derivative" is actually unsettled, Creative Commons licenses were written for creative works rather than model weights, and no court has drawn the line. This study offers no legal opinion, and none of its argument requires one.

The web demo illustrates how quickly the ambiguity becomes practical. Its documentation states that it runs the released checkpoints unmodified, phrasing evidently chosen with the license in view. The code converts the models to half precision and ports logic from the research repository. Under ordinary engineering usage, neither modifies the model; under a literal reading of the license, both are arguable. A careful developer acting in good faith ended up in a grey zone. No clearer evidence exists that the terms are hard to apply. A related tension appears on the dataset card. It lists intended uses including text-to-speech training and voice cloning, while the license prohibits distributing modified versions. Training a model on the data and releasing it is, plainly read, distributing something adapted from it. The card invites a use whose permissibility under its own terms is unclear, and leaves each user to resolve that alone.

**What would close this.** The researchers cannot unilaterally relax terms they inherited. But they are in a position to ask whether the audiobook rights holders would permit corrected or dialect-adapted versions of the models to be redistributed for non-commercial research, even with the recordings remaining under the present terms. A narrow model-only exception would let the community fix errors without touching the underlying content rights. Failing that, a short note on the dataset card clarifying which of the listed intended uses the license actually permits would spare every downstream user the same analysis.

In documentation terms, two small moves would help. First, state license provenance plainly: which terms come from the audiobook permissions, which from the F5-TTS weights, and which (if any) were chosen by the authors. Second, treat recordings, transcripts, model weights and code as separate artifacts with their own licenses, rather than applying one blanket CC BY-NC-ND across the release. That would preserve the community's ability to build on the result even when the raw recordings must stay restricted.

**9. One rights holder named, one absent**

The acknowledgments are specific about permissions:

> *"We sincerely thank Malaka Mustafa Soltani, the writer of the female audiobook, for kindly granting us permission to use the content of her audiobook. We also thank Shahin Shahlai, the narrator of the female audiobook, and Fateh, the narrator of the male audiobook, for allowing us to use their voices"* (p. 671).

For the female audiobook this is a complete chain. The author of the text granted permission for the content; the narrator granted permission for her voice.

For the male audiobook the chain is incomplete. The narrator is credited. No writer, publisher or content rights holder appears anywhere in the paper. Table 1 records that dataset as 10 hours 51 minutes and 6,044 utterances in the biography domain, a full-length work whose transcripts are reproduced in the public dataset viewer. Section 8 refers to *"the writers and narrators"* in the plural, but only one writer is named.

Permission may well have been obtained and simply not recorded in the paper. But a reader cannot determine who holds the rights to that text, and neither can anyone building on the corpus.

A minor note for anyone cross-referencing sources: the paper spells the narrators Shahin Shahlai and Fateh, while the dataset card spells them Shahen Shahlai and Fatih.

**What would close this.** One line in the dataset card naming the rights holder for the male audiobook content and the scope of the permission granted, matching the detail already provided for the female audiobook.

**10. The pronunciation decisions embedded in the models**

Two preprocessing decisions shape everything these models learned about Kurdish, and neither is discussed in the paper.

The grapheme-to-phoneme step is described in a single sentence citing Mahmudi and Veisi 2021. That underlying work, the academic basis for the AsoSoft converter, observes that Kurdish orthography has been contested for decades, identifying the short vowel /i/ and two specific letters as unresolved cases. Sorani script does not write that short vowel, so the converter must decide where it belongs. Every such decision encodes a position on how Kurdish ought to sound, and because the models were trained on the converter's output, those positions are now fixed in the weights.

The cumulative effect is a particular kind of speech. Every input passes through a rule system that resolves each ambiguity the same way. The models then learned from audiobook narration and studio dubbing, already the most deliberate registers Kurdish is spoken in. Change the preceding same way, and to same way. The output is therefore standardized twice over, once by the recording register, once by the converter. It is consistent, it follows written orthography closely, and it does not vary the way a speaker does. Nothing in the paper claims otherwise. But a developer hearing the demo will hear fluent, natural Kurdish and reasonably conclude the models have learned Kurdish pronunciation, when what they have learned is one rule system's account of it, delivered in one register.

The released code passes three settings to the normalizer explicitly, 'changeInitialR=True', 'deepUnicodeCorrectios=True' and 'additionalUnicodeCorrections=True', all matching the library defaults, and calls the G2P converter with no arguments at all. AsoSoft's README names the parameter without describing its behaviour; the substitution of word-initial ر with ڕ is a recognised Central Kurdish normalization convention, implemented through a replacement list in Normalize.py. It is a convention rather than a hidden assumption, but not a documented one.

A second default, 'singleOutputPerWord=True', is harder to assess. The name suggests the converter returns one pronunciation per word rather than several candidates. Ambiguous words would then be resolved to a single choice. AsoSoft's documentation lists the parameter without describing its behavior, but the source settles it: word_G2P in G2P.py returns the first of several ¶-separated candidate pronunciations when the flag is true, discarding the rest.

The specific losses are the ones a Kurdish-speaking listener would notice and a non-speaking developer would not. Word-initial ر becomes ڕ in every case. That is the documented convention and correct for most words, but the setting admits no exception for loanwords or for speakers who do not make the substitution. The unwritten short vowel is inserted wherever the rules place it, though the paper's own cited source identifies this as one of the unresolved cases in Kurdish orthography. Because singleOutputPerWord discards all but the first candidate, words with more than one accepted pronunciation reach the model with one, and the alternatives leave no trace in the training data, so no amount of prompting recovers them. These are not errors. They are choices, made consistently, that a listener from a different region may hear as slightly wrong in ways they could not articulate and a developer could not diagnose.

One choice can be established exactly. The converter marks syllable boundaries in its output, its own worked example reads '*ˈşeˈwû ˈřoj ˈbûyn ˈbe ˈgiˈrift*'. The pipeline strips every marker before the text reaches the model, in a single call chained onto the converter's output: KurdishG2P(norm).replace("ˈ", ""). What makes this notable is that the marker symbol is present in 'vocab.txt', so it was not removed out of necessity. Whether retaining it would have improved results is not something this study can determine. The paper identifies prosodic variability as the most challenging case (p. 668), with MOS of 3.88, 4.01 and 3.95 (Table 3, p. 669), though the expression category scores lower still for all three systems (3.66, 3.61, 3.89), while acknowledging that the two facts are not necessarily connected.

**What would close this.** A short statement of which converter settings were used, that syllable information is discarded before training, and that the models were trained on read speech would let a reader calibrate what they are hearing. That last point is the one most easily assumed away: an English-language developer evaluating this release will apply intuitions formed on systems trained across many speakers and registers, where a single model plausibly represents the language. Here it represents one narrator, one register and one rule system, and the paper's evaluation supports it on exactly those terms.

For a linguist the practical question is different and sharper. Documentation and revitalization work usually targets spoken varieties rather than the written standard, and this pipeline normalizes toward the standard by design before any model is trained, before any output is judged. A researcher who records regional speech and pushes the transcripts through this preprocessing chain will get audio that sounds like standard Central Kurdish reading their words back, not their variety being spoken. The pipeline cannot preserve what it was built to remove. Whether that is a problem depends entirely on the purpose, and knowing it in advance requires reading a dependency's documentation that the paper cites in one line and the repository does not mention at all.

## 11. The direction of accommodation

Each preprocessing decision in this pipeline is defensible on its own terms. What the paper does not record is that they run in the same direction. Taken in sequence, the pipeline removes linguistic variation at every stage, and the removal is cumulative, so that what reaches the model has been standardized several times over and what the model can produce is bounded by the narrowest filter in the chain.

The stages are these, and all are documented either in the paper or in the released files.

**Source register:** All three corpora are read speech: audiobook narration in the biography domain, and a professional dubber reading prepared sentences (Table 1, p. 665). Published audiobooks are edited toward a literary norm before anyone records them. Regional phonology, local vocabulary, code-switching and spontaneous prosody are substantially gone before the first engineering decision is made.

**Speakers:** Three, with variety and region unrecorded. Whatever those three speak constitutes the entire attested range of the corpus, and no reader can determine what that range is.

**Transcription:** Utterances were transcribed automatically by a fine-tuned recognizer and then manually revised (p. 665). The recognizer's spelling conventions therefore shape the text before any human sees it, and the orthographic norms applied during revision are not reported.

**Normalization.** The AsoSoft normalizer is called with changeInitialR=True, converting word-initial ر to ڕ in every case. The substitution is a recognized Central Kurdish convention and correct for most words, but the setting admits no exception for loanwords or for speakers who do not make it.

**Grapheme-to-phoneme conversion.** The converter resolves the cases that its own academic basis, Mahmudi and Veisi (2021), identifies as long contested in Kurdish orthography: the unwritten short vowel and the letters ی and و. It resolves them the same way every time. singleOutputPerWord=True then discards all candidate pronunciations after the first, so words with more than one accepted realization reach the model with one, and the alternatives leave no trace in the training data.

**Vocabulary:** The resulting phoneme string must land in a 2,545-entry inventory containing pinyin with tone numbers, hangul, kana, Cyrillic, Greek, Hebrew, Thai and extended Latin, and nothing Kurdish-specific.

**Evaluation:** The raters' varieties are unrecorded, so naturalness was measured against an audience of unknown composition. The recognizer scoring intelligibility is the one that produced the transcripts, so it is best calibrated to the conventions the pipeline already imposed.

**License:** CC BY-NC-ND forbids redistribution of modified versions. No one outside the project can publish a system that reverses any decision above.

**Which party adapted**

The sixth stage matters most, because the paper's own sentence records something the discussion around it does not draw out. The conversion to phonemes is applied *"to make it compatible with the vocabulary of the F5-TTS baseline model"* (p. 666). Read plainly, that is a statement about direction. The vocabulary was fixed by a checkpoint trained on a large Chinese-and-English corpus. Kurdish was converted until it fitted.

The alternative was available in principle: extend the vocabulary with the characters Kurdish requires and continue pretraining. It was not taken, almost certainly because it costs data and compute this project did not have, and that is a reasonable decision under constraint. But it should be recorded as a decision, because its consequence is durable. The representational space available to Kurdish in these models is a space that was never designed with Kurdish in it. Anything Kurdish that has no counterpart in that inventory is not merely modelled poorly. It cannot be represented at all.

Two instances are already visible in the release. The converter emits the Hawar Latin characters ḧ and ẍ for ح and غ. Neither appears in vocab.txt. So is <, which is why the number placeholder described in Section 6 fails. Neither absence is signaled anywhere, and the pipeline contains no step that reports characters it cannot represent, so the failure mode is silence rather than error.

The justification offered for the whole arrangement is that Central Kurdish is *"almost a phonemic language and have very high precision Grapheme-to-Phoneme (G2P) tools"* (p. 666). That claim carries the design, and it is supported by no citation. The paper's own dependency reports contestation in exactly the places where a phonemic claim would need to hold. The word doing the work in that sentence is *almost*, and *almost* is where the variation lives.

**Parallel standards, not a standard and its deviations**

The paper's opening establishes that Kurdish is a macrolanguage spanning a dialect continuum across more than 35 million speakers (p. 664). What that framing understates is that the written situation is not a single standard with regional departures from it. Sorani is written in modified Arabic script; Kurmanji in Hawar Latin in Turkey and Syria and in Arabic script in Iraq; Kurdish in the former Soviet republics has historically been written in Cyrillic. These are parallel standards, maintained by different communities under different states, and no institution has the authority to rank them.

That condition is what makes the normalization question sharper here than it would be for a language with an academy. A pipeline that resolves contested orthographic cases by rule is not selecting between a correct form and an error. It is selecting among the positions of communities that have not agreed, and it is doing so through a software default that neither the paper nor the repository names.

The consequence for reuse is concrete. The paper offers a method for rapid development of synthesis in low-resource settings, and the same group has announced multi-dialect resources extending to Badini and Hawrami. A researcher applying this recipe to either inherits the script and orthographic assumptions along with it, because those assumptions live in a dependency rather than in the code, and nothing in the release tells them the assumptions exist.

**What makes a high-resource language high-resource**

The comparison this release invites is with English and Chinese systems, since it is fine-tuned from English-and-Chinese weights and assessed with metrics developed on such languages. That comparison conceals an asymmetry which has nothing to do with hours of audio.

Synthesis in English or German works as well as it does partly because of accumulated descriptive work: orthographies with institutional backing, pronunciation dictionaries compiled and revised by phoneticians across generations, phonological descriptions, dialect surveys, lexicographic traditions, and corpora annotated by people trained to annotate them. Mandarin has a state-backed romanization and a philological tradition older still. None of this is a dataset in the machine-learning sense and none of it appears on a model card. It is the layer underneath, and it is what makes an automated decision auditable. When a converter produces an unexpected pronunciation for an English word, there is a dictionary to check it against, a description of the relevant process, and usually a record of which varieties exhibit it.

High-resource corpora are not hand-annotated while low-resource ones are automated. The checkpoint these models are fine-tuned from was itself pretrained on a large in-the-wild corpus prepared by an automated pipeline, and English systems make thousands of unexamined decisions of exactly this kind. What differs is whether the automation can be checked. When an English pipeline resolves an ambiguity wrongly, somebody can establish that it did. For Central Kurdish there is no comparable reference, so the converter's decisions are not merely unexamined; within the terms of this release they are unexaminable.

That is the situation in which the paper describes Central Kurdish as *"almost a phonemic language and have very high precision Grapheme-to-Phoneme (G2P) tools"* (p. 666), with no citation attached. The claim may well be correct. But it is the kind of claim that a body of descriptive work exists to settle, and here the automation stands in for the settlement rather than resting on it.

A further asymmetry sits in the pipeline itself. The paper notes that F5-TTS ordinarily avoids grapheme-to-phoneme conversion (p. 666), working from text directly. English and Chinese therefore reach the model with one fewer irreversible transformation than Kurdish does, and the Kurdish transformation was introduced for vocabulary compatibility rather than for any property of the language. The languages with the deepest descriptive resources require the least preprocessing. The language with the fewest acquires an additional step, applied by a tool whose own source reports the underlying questions as open.

This is where the paper's citation practice becomes a substantive matter rather than a formal one. Kurdish linguistics is not an empty field: there is dialectological and descriptive work on the varieties, work on standardization and language policy, and work on Sorani phonology and orthography, including the Optimality Theory analysis from which the paper's own converter derives and which it cites once, for the tool rather than for the analysis. Of roughly thirty-six references, two are linguistic, and both appear in adjacent sentences of the introduction.

Citing that work would not be a courtesy. It does three things the release currently cannot do for itself. It tells a reader which of the pipeline's linguistic commitments are settled and which are contested. That is what a user needs in order to interpret the output. It gives a researcher outside Kurdish studies a route into the field. The paper's stated audience needs that route in order to judge whether the method transfers to their own language. And it locates the release within a body of scholarship on which the eventual improvement of Kurdish speech technology depends, rather than presenting the engineering as though it could proceed independently of it.

The constructive form of the argument is this. English has good pronunciation resources because people did the descriptive work over a long period and recorded it so that others could check it. Kurdish speech technology will need the same, and will need it more urgently, because the varieties with no coverage are those where the descriptive gap is widest. A release that treats grapheme-to-phoneme conversion as a solved dependency passes over the point at which that collaboration is most needed. The authors are well placed to name it, and naming it costs a paragraph.

**How a default becomes a fixture**

No one is compelled to use these models. The mechanism is availability, and it is slower and more durable than compulsion.

Where a language has one working system and no alternative, that system becomes the reference point by default. Applications are built on it because there is nothing else to build on. Later systems are benchmarked against it because it is the only prior result. Listeners who encounter synthetic Kurdish for the first time encounter it there, and what they hear establishes what synthetic Kurdish sounds like. The model card recommends the female audiobook system as *"suitable for general-purpose applications"*, and general purpose, for a language of roughly eight million speakers whose regional variation the release does not record, is a considerable extension of what was demonstrated.

The no-derivatives term is what converts the default into a fixture. A developer in a region whose variety realizes a vowel differently can hear the discrepancy, can correct it privately, and cannot publish the correction. The restriction originates with the audiobook rights holders rather than with the researchers, but its effect on the next developer does not depend on where it came from.

**What this criticism is not**

The authors are not indifferent to variation. Their subjective evaluation includes categories for code-switching, idiom and prosodic variability precisely to probe beyond the training register, and it reports prosody as the hardest case (p. 668). They state directly that three speakers are insufficient for a multi-speaker model (p. 666). They are separately building a multi-dialect recognition benchmark covering Central, Northern and Southern Kurdish, with Hawrami and Badini announced.

Nor is fine-tuning from a multilingual checkpoint the wrong choice. It is what makes eleven hours of audio sufficient, and no Kurdish variety has the data to train a system of this quality from scratch. Borrowing pretrained weights, and with them the vocabulary those weights were trained against, is what low-resource work consists of at present.

The criticism is narrower and, for that reason, harder to set aside. Standardization is a property of this pipeline rather than a limitation the authors failed to overcome, and a property of that magnitude should be visible in the artifacts. As released, a user cannot tell how much of what they hear is Central Kurdish and how much is one rule system's account of it delivered in one narrator's reading register. The paper's evaluation supports the models on exactly those terms. The model card does not say so.

**What would close this**

None of the following requires additional data or another training run.

Record speaker variety and region in the dataset card, and the dialect distribution of the 88 raters. Both are information the team already holds, and their absence means no one can tell whether the 0.02 margin separating the top two systems reflects the systems or reflects which variety the listeners happened to speak.

State that the pipeline normalizes toward the written standard, at which stages this occurs, and that the models were trained on read speech. Report which converter settings were used and that syllable markers are discarded before training.
Add the vocabulary-coverage check proposed in Section 6, reporting characters absent from the inventory before synthesis. It costs a single lookup and converts every failure of this class from silent to visible. That matters most for anyone extending the pipeline to a variety written in another script.
Describe the release for what the evaluation supports: Central Kurdish read-speech synthesis, one speaker per voice, rather than general-purpose Kurdish synthesis.
Ground the linguistic claims that carry the design in the relevant scholarship, particularly the description of Central Kurdish as almost phonemic and the treatment of the converter's output as an adequate phonemic representation for training. A handful of references establishing what is settled about Sorani phonology, what remains contested in the orthography, and how the varieties are conventionally divided would let a reader outside Kurdish studies evaluate the method rather than take it on trust.
Ask the audiobook rights holders whether dialect-adapted models may be redistributed for non-commercial research while the recordings remain restricted. Of everything in this study, that is the item whose resolution determines whether the second Kurdish variety to acquire a voice can build on the first.

**12. Which variety, and whose judgement**

The paper opens by describing Kurdish as a macrolanguage spanning a dialect continuum, Northern, Central, Southern, Hawrami, Zazaki and Laki, across more than 35 million speakers, with Central Kurdish at approximately 8 million (p. 664, citing Sheyholislami 2015 and 2021).
Having established that framing, the release records nothing further about variety. Table 1 lists duration, utterance count and domain for each speaker, but not which variety of Central Kurdish they speak. The subjective evaluation reports 88 participants and 3,101 answers without describing their linguistic backgrounds.
This is plainly not indifference to the question. The same account publishes a Kurdish multi-dialect speech recognition benchmark whose current version includes Central Kurdish, Northern Kurdish and Southern Kurdish; its dataset page states that Hawrami and the Badini variant of Northern Kurdish will be added soon. The group works on dialect coverage actively.
The framing itself rests on a thin foundation, and this is worth stating precisely rather than as a general complaint. The paper's bibliography runs to 36 references, of which two are linguistic: Sheyholislami 2015 and 2021, both cited in adjacent sentences of the introduction to establish that Kurdish is a dialect continuum and that Central Kurdish has around 8 million speakers. The remaining citations to Kurdish-specific scholarship are to tools: Mahmudi and Veisi's G2P work and the AsoSoft normalization paper. Both are engineering resources that happen to concern Kurdish rather than descriptions of the language. There is no reference to work on Sorani phonology beyond the G2P paper itself, or on regional variation within Central Kurdish. Sheyholislami 2021, on the history and development of literary Central Kurdish, is the only citation touching standardization, and it is cited once, for a speaker figure, or on the sociolinguistics of standardization in Kurdish. Mahmudi and Veisi 2021 is the exception, an Optimality Theory analysis that does engage the orthographic questions, but it is cited once, for the tool it produced rather than for the analysis behind it.
This is not a linguistics paper and should not be judged as one. But it does make claims that are linguistic in kind: that Central Kurdish is "almost a phonemic language" (p. 666), that its varieties form a continuum, that a converter's output is an adequate phonemic representation for training. Those claims carry the design. Grounding them in the literature is ordinary scholarly practice, and here it would also do practical work.
The practical work is this. A reader outside Kurdish studies has no route from this paper into the field. They cannot check whether "almost phonemic" is a settled description or a working simplification; whether the varieties named are conventionally divided that way or whether the classification is disputed; whether the orthographic uncertainties the converter resolves are minor or central. The paper's own dependency identifies the short vowel /i/ and the letters ی and و as long-contested. That points to the second answer, but a reader would have to follow the citation two steps to discover it. For non-Kurdish researchers evaluating whether the method transfers to their own languages, the audience the paper explicitly addresses, the absence of that grounding is the difference between an informed judgement and a guess.

Returning to variety: this release does not record it, and two groups of readers need that information. A developer selecting a voice for an application wants to know how it will sound to the people who will hear it. A researcher reading the listening test wants to know whether raters shared a variety with the speakers, since that shapes how the scores should be interpreted, especially where an entire ranking rests on 0.02.

Register is missing too, and for Kurdish it may matter more than variety. All three corpora are read speech: audiobook narration in the biography domain and a professional dubber reading prepared sentences (Table 1, p. 665). Read speech is deliberate, consistently paced, and follows written orthography. It is not how the language is spoken in a classroom, a market, or a phone call, and spontaneous Kurdish carries regional phonology, local vocabulary, code-switching and prosodic patterns that written registers smooth away before a recording is ever made.

The paper is aware of the limit and tests for it. Its subjective evaluation includes categories for code-switching, idiomatic expressions and prosodic variability precisely to probe beyond the training register, and reports that prosodic variability was the most challenging case (p. 668), with MOS of 3.88, 4.01 and 3.95 (Table 3, p. 669). That is a candid result, and it marks the boundary of what read-speech training supports.

The consequence for extending this work to other varieties is practical. A researcher hoping to build a Badini or Hawrami system will need to establish whether suitable speech resources exist, rather than assuming that the Central Kurdish audiobook-based recipe transfers directly to those varieties. The recipe demonstrated here depends on a substantial, prepared speech resource, so its transferability to varieties without comparable material remains an open question. The finding that audiobook data works should not be read as a general solution for varieties that have no audiobooks.

There is a longer-term consideration, best stated as a risk rather than a prediction. Where public resources are scarce, whatever exists tends to become the reference point, not by anyone's decision but by default. A reference model carries its choices forward: the converter's phonemic rules, the transcription conventions, the voices of three speakers whose varieties are recorded nowhere. Because the license prevents redistribution of modified versions, a researcher who identifies a mispronunciation or wants to adapt a voice to another variety cannot publish the correction. The restriction originates with the rights holders rather than the researchers, but its effect on the next developer is identical.

**What would close this.** Two columns in the dataset card recording speaker variety and region, and one line reporting the dialect distribution of the 88 raters. Both are information the team already holds, and their absence has a specific cost: without them, no one can tell whether the 0.02 margin separating the top two systems reflects the systems or reflects which variety the listeners happened to speak. Regional differences in Kurdish are audible, and the paper does not report where the raters were recruited.

A short paragraph situating the work in Kurdish linguistics would close the second gap, and it need not be long: a handful of references establishing what is settled about Sorani phonology, what remains contested in the orthography, and how the varieties are conventionally divided. For a paper whose stated audience is researchers working on other low-resource languages, that paragraph is what allows them to judge whether their own language resembles this one closely enough for the method to carry over. Without it, the recipe is offered without the conditions under which it applies.

### 13. What the release does well

Closing on gaps alone would misstate what is here, so it is worth being precise about what the release does establish and on what basis.

Table 3 gives 95% confidence intervals for every metric across seven categories and three systems. Those intervals are what allowed the 0.02 margin in Section 4 to be put in proportion. Table 5 supplies a zero-shot translation baseline of 2.31 and 1.51 BLEU against the 27.23 and 18.50 obtained with synthetic training data, and Table 4 an upper bound of 92.42 from actual FLEURS audio, without those figures the headline numbers would be uninterpretable. The evaluation code is published, and it runs: the bundled test configuration reproduced its own shipped reference values across all seven metrics. Much of the analysis in this study was possible only because the paper reports its own results at this level of detail.
What that reporting can support is bounded by the data behind it. Each condition rests on one speaker recorded reading prepared material. The paper acknowledges this when it explains that the speaker count was insufficient for a multi-speaker model (p. 666). Confidence intervals computed over categories within a single speaker describe variation across sentence types, not across speakers or varieties or registers. The evaluation is therefore precise about a narrow question, how three specific systems compare on read speech, judged by 88 listeners whose backgrounds are unrecorded, and the paper's language stays within that boundary in a way the model card's does not.
The corpus is the most durable part of the release. Thirty-five hours of transcribed Central Kurdish speech, with permission recorded from a named author and both narrators, is a resource that outlasts any particular model trained on it, and it remains useful for purposes the models are not, including recognition work and phonetic study. Its limits are the ones described throughout: three speakers, read register, undocumented varieties. The Giganet studio dataset had, to the authors' knowledge, never been systematically evaluated or used to build a text-to-speech model before this work (p. 665), so bringing it into a comparison is itself a contribution.
The independently built web demo. It displays the phonetic string the model receives. That is good practice, and it is how the preprocessing behavior examined above can be inspected at all. It also declines user-uploaded reference audio, so it cannot become a general-purpose tool for cloning arbitrary voices.

## 14. Conclusion

A pattern runs through this study. The paper reports its work carefully and states its limits, and the artifacts through which most people will encounter that work do neither. A model card asserts a firmer conclusion than the paper draws. A configuration file describes hardware that was not used. A dataset omits the split its own evaluation depends on. A preprocessing step appears in the code, not in the paper, and does not do what it was written to do.
Underneath those specific mismatches sits a broader one. What was demonstrated is narrow: three single-speaker systems trained on read material, evaluated by listeners whose varieties were not recorded, with the ranking between them resting on two hundredths of a point. What the release presents is more general, a method for rapid development of speech synthesis in low-resource settings, offered to researchers working on other languages. The distance between the two is where most of the difficulties in this study are found, and none of them require the research to be wrong. They require only that a reader take the artifacts at face value, which is what artifacts are for.
That distance costs more for Kurdish than it would elsewhere. A well-resourced language has alternatives, so an incomplete release is one option among several. Central Kurdish has very few, and the varieties with least coverage depend on exactly the researchers least equipped to reconstruct what the documentation omits. A missing training command or an unmarked test split is a minor inconvenience to an engineer and a wall to a linguist, and it is linguists who do most documentation work on the varieties that still have none.
Much of this is cheap to fix, and the specific remedies are listed at the end of each section. Marking the test split is one column. Aligning the model card with the paper's own wording is three sentences. Correcting the configuration is two lines. Publishing the evaluation sentences involves no third-party rights. Recording speaker varieties and rater backgrounds uses information the team already holds. None of these changes any result; they change whether the results can be checked.
The license is the one item the researchers cannot resolve alone, since the terms came from the audiobook rights holders. The constructive route runs through those rights holders: whether permission might be sought for corrected or dialect-adapted *models* to be redistributed for non-commercial research, while the *recordings* remain under existing terms. Whether that permission is available determines something larger than this release, whether the next Kurdish variety to acquire a voice can build on this work, or has to begin again.

**The following are unresolved in the public materials. Each affects how the release can be used or verified, and most could be settled by information the authors already hold.**

**On the training run**

1. Which hardware configuration produced the released checkpoints. The configuration file carries an eight-GPU comment and specifies eleven epochs; the paper reports one GPU, and Figure 1 extends to approximately 275,000 steps.
2. Which checkpoint among those steps was released, and on what basis. Section 4 opens by noting that no standard metric determines when to stop training, and the selection is not reported.
3. The TTS training command, the data manifest format, and any audio preparation applied beyond the resampling noted on page 665. The recognizer's fine-tuning script is published; the TTS equivalent is not.

**On the evaluation**

4. Whether the 1,500 held-out test utterances can be identified within the released corpus. As distributed, anyone fine-tuning on this data trains on the paper's evaluation set.
5. Whether the seven categories of evaluation sentences can be published, without which the listening test cannot be repeated.
6. How the cross-system character error rates should be read, given that the recognition system used to score them is the one that generated the transcripts on which the models were trained and shows a markedly higher error rate on genuine studio speech than on genuine audiobook speech (0.09 against 0.01 and 0.02, with a wide interval on the studio figure).
7. Whether the Central Kurdish recognizer behind the intelligibility results can be released without an access barrier. Its fine-tuning script is published in the evaluation repository; a file listing for the corresponding model on Hugging Face returned HTTP 401 when requested without credentials.

**On language and documentation**

8. Which varieties of Central Kurdish the three speakers represent, and the dialect backgrounds of the 88 raters.
9. Which grapheme-to-phoneme settings were used, and whether the removal of syllable markers was deliberate or an inherited default.

**On rights and reuse**

10. Who holds the rights to the text of the male audiobook, and the scope of the permission granted.
11. Whether the model card can be brought into line with the paper, recording the starting checkpoint, and using the paper's own terms "competitive" and "marginally better."
12. Whether the audiobook rights holders would permit redistribution of corrected or adapted models for non-commercial research, with the recordings remaining under existing terms.

**Model**

https://huggingface.co/aranemini/central-kurdish-tts. Model card text and repository files (F5TTS_v1_Base.yaml, vocab.txt, infer.py) retrieved 08.09.2026 at commit cdca52a. All model card quotations in this study are from that revision.

**Dataset**

https://huggingface.co/datasets/aranemini/central-kurdish-tts4all. Dataset card and viewer row count retrieved 08.09.2026.

Related resources by the same account. aranemini/kurdish-multidialect-asr-benchmark and aranemini/frezar; the dataset page describes the current version as covering Central Kurdish, Northern Kurdish and Southern Kurdish, with Hawrami and the Badini variant of Northern Kurdish announced as forthcoming. https://huggingface.co/datasets/aranemini/kurdish-multidialect-asr-benchmark and https://huggingface.co/datasets/aranemini/frezar. Retrieved 08.09.2026.

**Demonstration**

https://huggingface.co/spaces/tebinraouf/central-kurdish-tts. Application code and documentation retrieved 08.09.2026 at commit b5c6a0a. Built independently of the research team.

**Upstream components**

F5-TTS: https://github.com/SWivid/F5-TTS. Code MIT; pretrained weights CC-BY-NC. Base checkpoint and training-data description from src/f5_tts/infer/SHARED.md; training and dataset preparation from src/f5_tts/train/README.md and src/f5_tts/train/datasets/prepare_csv_wavs.py. Retrieved 8 September 2026 at commit 9c614e9.

**Evaluation code**

TTS4ALL_Eval, https://git-lium.univ-lemans.fr/jsalt2025/wp1/tts4all_eval (cited at p. 667), commit 116761d, dated 29 May 2026. GNU AGPL v3.0. Retrieved 8 September 2026.

**Not established**

The identity of the literary works in either audiobook; the rights holder for the male audiobook text; whether the 'aranemini' account belongs to the paper's lead author. No claim in this study depends on any of these.